\documentclass{article}
\usepackage{arxiv}
\usepackage[utf8]{inputenc}

\usepackage{verbatim}
\usepackage{ifthen}
\usepackage{xspace}
\usepackage{pgf}

\usepackage{amsmath,amssymb}

\usepackage{url}

\usepackage[vlined,ruled]{algorithm2e}
\usepackage{tikz}

\usepackage{color}
\usepackage{xcolor}
\usepackage{colortbl}
\usepackage{tcolorbox}

\usepackage{ucs}
\usepackage{comment}
\usepackage{amsmath}
\usepackage{fancybox}
\usepackage{caption}
\usepackage{subcaption}
\usepackage{bm}
\usepackage{rotating}

\usepackage{array}
\usepackage{multirow}

\usepackage{xspace}
\usepackage{hyperref}

\newcommand{\burl}[1]{\structure{\url{#1}}}

\newcommand{\rl}{RL\xspace}

\def\bbbe{{\rm I\!E}} 

\newcommand{\E}{{\bbbe}{}}

\newcommand\no[1]{}
\newcommand\wi[1]{$\circ$}
\newcommand\bu[1]{$\bullet$}
\newcommand\ot[1]{$\star$}
\newcommand\bo[1]{$\bullet\star$}

\newcommand{\mujoco}{{\sc mujoco}\xspace}
\newcommand{\openai}{{\sc openAI}\xspace}
\newcommand{\gym}{{\sc gym}\xspace}
\newcommand{\swimmer}{{\sc swimmer}\xspace}

\newcommand{\trpo}{{\sc trpo}\xspace}
\newcommand{\tqc}{{\sc tqc}\xspace}

\newcommand{\sac}{{\sc sac}\xspace}
\newcommand{\tddd}{{\sc td}3\xspace}

\newcommand{\cem}{{\sc cem}\xspace}

\newcommand{\rien}[1]{}

\DeclareGraphicsExtensions{.jpg,.mps,.pdf,.png,.gif}

\definecolor{myred}{rgb}{0.8,0,0}
\definecolor{mygreen}{rgb}{0,0.6,0}
\definecolor{myblue}{rgb}{0,0,0.7}

\newcounter{ques} 
\newcommand{\ques}{\arabic{ques}}

\title{Making Reinforcement Learning Work on Swimmer}
\author{Ma\"{e}l Franceschetti$^{(1)*}$, Coline Lacoux$^{(1)*}$, Ryan Ohouens$^{(1)*}$,
\href{https://orcid.org/0000-0001-6036-6950}{\includegraphics[scale=0.06]{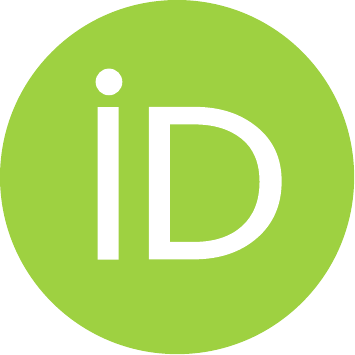}\hspace{1mm}{\bf Antonin Raffin}$^{(2)}$}\\
and \href{https://orcid.org/0000-0002-8544-0229}{\includegraphics[scale=0.06]{orcid.pdf}\hspace{1mm}{\bf Olivier Sigaud}$^{(1)}$},
\\
(1) Sorbonne Universit\'e, CNRS, Institut des Syst\`emes Intelligents et de Robotique,\\ F-75005 Paris, France\\
(2) German Aerospace Center (DLR), Robotics and Mechatronics Center (RMC),\\ M\"unchner Str. 20, 82234 We\ss ling, Germany\\
{\em correspondence to} Olivier.Sigaud@isir.upmc.fr\\
$^*$: equal contribution}

\begin{document}

\maketitle

\begin{abstract}
The \swimmer environment is a standard benchmark in reinforcement learning (\rl). In particular, it is often used in papers comparing or combining \rl methods with direct policy search methods such as genetic algorithms or evolution strategies.
A lot of these papers report poor performance on \swimmer from \rl methods and much better performance from direct policy search methods. In this technical report we show that the low performance of \rl methods on \swimmer simply comes from the inadequate tuning of an important hyper-parameter, the discount factor. Furthermore we show that, by setting this hyper-parameter to a correct value, the issue can be easily fixed. Finally, for a set of often used \rl algorithms, we provide a set of successful hyper-parameters obtained  with the Stable Baselines3 library and its \rl Zoo.
\end{abstract}

\section*{Introduction}

A current trend in reinforcement learning (\rl) research consists in evaluating newly proposed algorithms on a large set of benchmarks to facilitate comparison with the existing literature.

In this context, the \swimmer environment is a standard \rl benchmark, it is used in a large number of papers to evaluate the performance of various reinforcement learning algorithms. In particular, it is often used in papers comparing or combining \rl methods with direct policy search methods such as genetic algorithms or evolution strategies.

A lot of these papers report poor performance on \swimmer from \rl methods and much better performance from direct policy search methods (see Section~\ref{sec:lit}). 

In this technical report, we investigate the reason behind this gap of performance. By analysing the behavior of the best solutions found by various methods, we show that the low performance of \rl methods on \swimmer simply comes from the inadequate tuning of an important hyper-parameter and that, by setting this hyper-parameter to a correct value, the issue can be very easily fixed.

From this analysis, we conclude that previously published results on \swimmer that do not use our simple fix should be regarded with care and, in particular, in all papers comparing or combining \rl methods with direct policy search methods, the conclusions drawn from the \swimmer benchmark should be reconsidered.

\section{The \swimmer environment}

The \swimmer environment is a standard reinforcement learning (\rl) benchmark built on the \mujoco simulation environment \cite{todorov2012mujoco} and accessed through the \openai~\gym interface \cite{brockman2016openai}.
The most used versions in the works listed in this paper are {\sc swimmer-v2} and  {\sc swimmer-v3}, but below we simply refer to it as \swimmer.

\begin{figure}[!htbp]
  \begin{center}
{\includegraphics[width=0.99\linewidth]{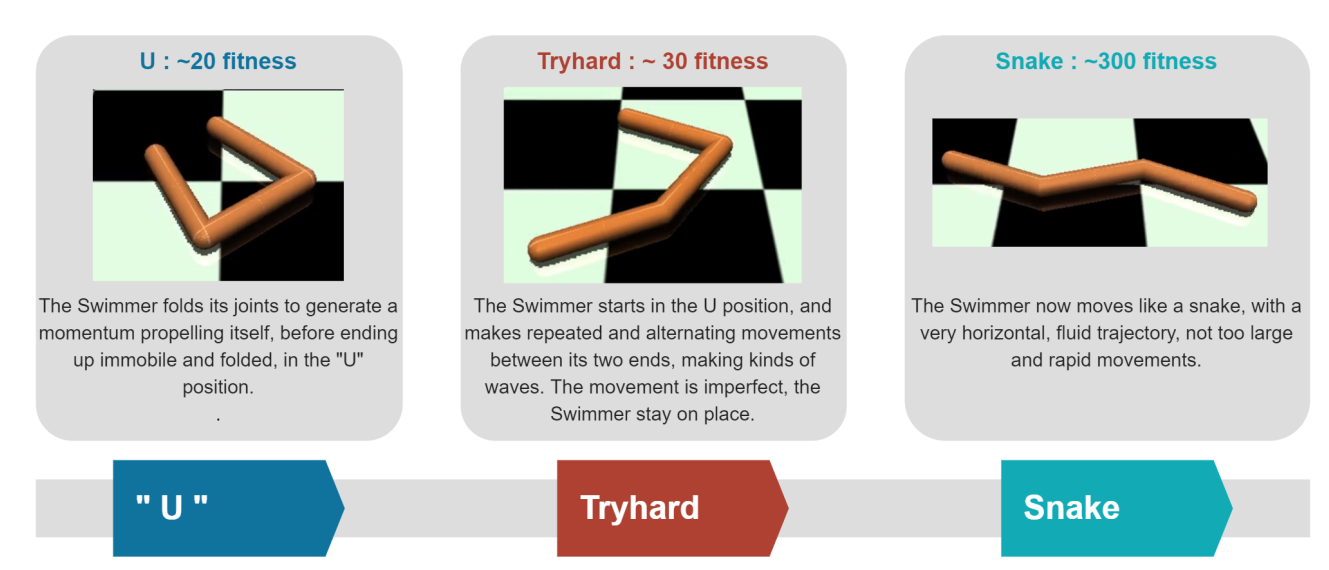}}
   \caption{
   Views of the \swimmer environment depicting three different control strategies with their approximate performance \label{fig:swimmer_view}}
   \end{center}
\end{figure}

In this environment, a planar creature made of three sticks must swim as fast as possible into a liquid. The policy controls the two degrees of freedom of the creature in rotation, namely the articulations between the sticks, by applying torques.
To simplify descriptions, we will call the most forward extremity of the creature the head, its most forward articulation the neck, and its most backward body the tail. Views of the environment can be found in \figurename~\ref{fig:swimmer_view}.

Beyond these static views, at the time we write this paper some videos of the swimmer in action are available on the following website: \href{https://72indexdescartes.wixsite.com/swimmer}{https://72indexdescartes.wixsite.com/swimmer}.

In \swimmer, the performance returned at each time step corresponds to the instantaneous velocity of the creature at that time step. This instantaneous velocity could have been measured at the center of mass of the body but, to simplify calculations, velocity at the neck is considered. The global performance is the sum of these instantaneous returns over 1000 time steps.

\section{Low performance of \rl in \swimmer: some literature}
\label{sec:lit}

The \swimmer environment is often used in papers comparing or combining \rl algorithms and direct policy search methods, e.g. \cite{khadka2018evolutionaryNIPS, pourchot2018cem,chang2018genetic,shi2019fidi,bodnar2020proximal,zheng2020cooperative,suri2020maximum,kim2020pgps,wang2022surrogate}. This list is certainly far from exhaustive and, as we shall demonstrate in the next section, results of the \rl part in these methods reported on \swimmer should be disregarded, as they result from improper tuning of the \rl algorithms they use. The same is also true of several pure \rl papers, such as \cite{srivastava2019training, shen2020deep, cheng2021heuristic}.

One of the reasons why \swimmer is so popular in methods comparing or combining \rl algorithms and direct policy search methods is the contrast between the performance of both approaches. With evolutionary methods such as the Cross-Entropy Method (\cem) \cite{mannor2003crossentropy, rubinstein04crossentropy, deboer05tutorial}, one can easily obtain a performance over 300. The best performance we found in the literature is 362, obtained using Augmented Random Search \cite{mania2018simple}. By contrast, performance barely reaches 100 with deep \rl methods, with a strong local minimum around 40.

\section{Explaining and fixing the issue}

In the previous section we have outlined a performance gap between \rl and direct policy search on \swimmer. The goal of this section is to understand the reasons behind this performance gap.

Why is it that some optimization approaches reach a performance over 300 whereas others plateau below 120? To answer this question, we first analyzed the best behaviors found with the different approaches.

\begin{figure}[!htbp]
  \begin{center}
{\includegraphics[width=0.99\linewidth]{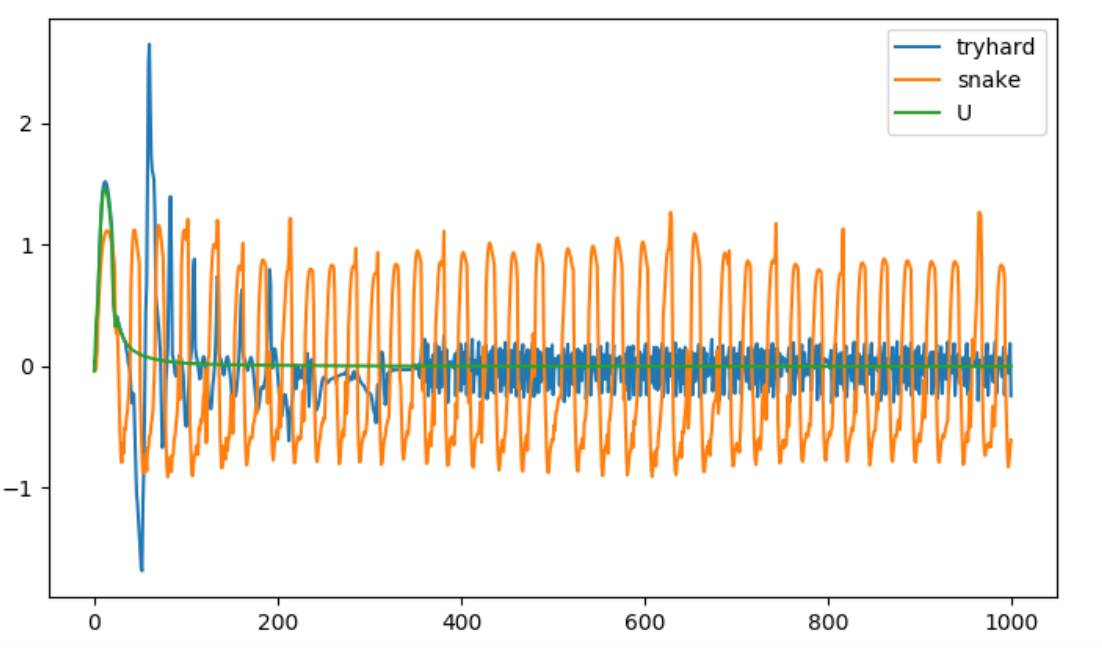}}
   \caption{
   Velocity of the \swimmer using the three different control strategies depicted in \figurename~\ref{fig:swimmer_view} \label{fig:swimmer_velo}}
   \end{center}
\end{figure}

To understand the global performance, \figurename~\ref{fig:swimmer_velo} depicts the velocity of the neck over 1000 time steps of three representative behaviors.

By comparing the curves, one can see that in two of the controllers, a peak velocity higher than that we can get with a snake-like behavior is reached in the very begin of the trajectory, whereas the velocity is much lower afterwards. In the case of the controller that moves the swimmer body into a U shape, the velocity quickly converges to 0 as the swimmer agent cannot generate further velocity from the configuration it reached.

Now, why is it that \rl methods inappropriately favor the first two types of behaviors whereas direct policy search methods find the better performing snake-like behavior?

The point is that in deep \rl, immediate rewards over a trajectory are aggregated using a discount factor $\gamma$, i.e. they optimize $J(\theta) = \E_{r_t \sim \pi_\theta}[\sum_t [\gamma^t r_t]]$, where $\pi_\theta$ is a policy parametrized by parameters $\theta$, e.g. a neural network.

In \rl, it is common practice to set $\gamma = 0.99, 0.95$ or even lower values. But discounting the reward changes the objective that the \rl agent is solving.

It happens that $0.99^{1000}=0.00004317124$, thus the last steps in reward aggregation are discounted by a factor $4.32\times 10^{-5}$ with respect to the first step. This is enough to explain that deep \rl agents maximize their velocity in the first steps and disregard what happens next.

\begin{figure}[!htbp]
  \begin{center}
  \subfloat[\label{fig:a1}]{\includegraphics[width=0.46\linewidth]{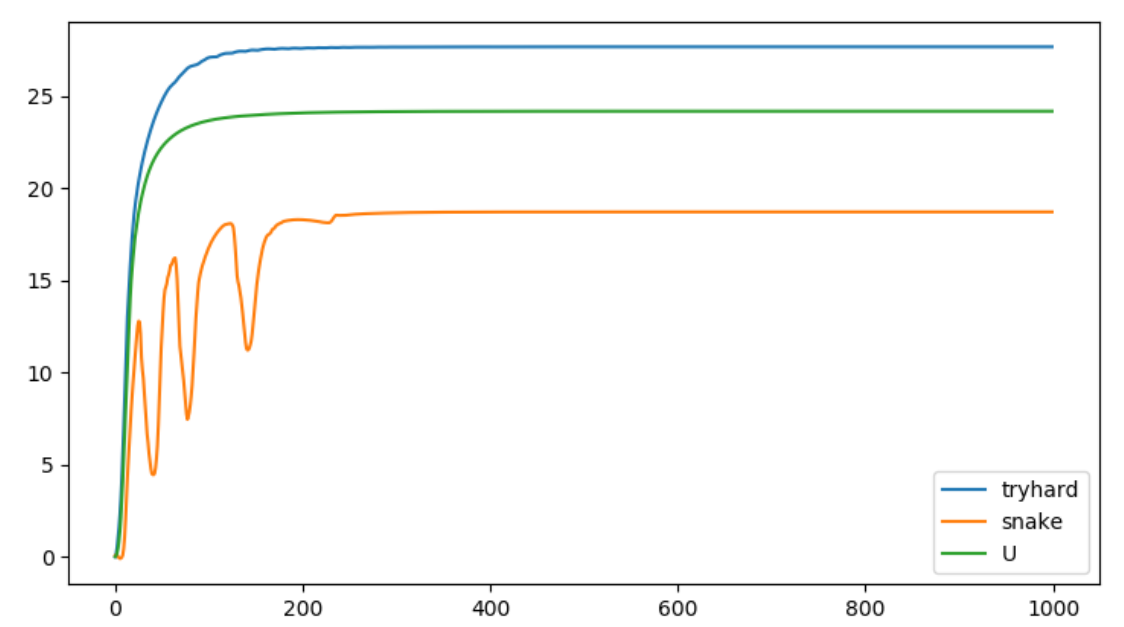}}
  \hspace{0.5cm}
  \subfloat[\label{fig:a2}]{\includegraphics[width=0.46\linewidth]{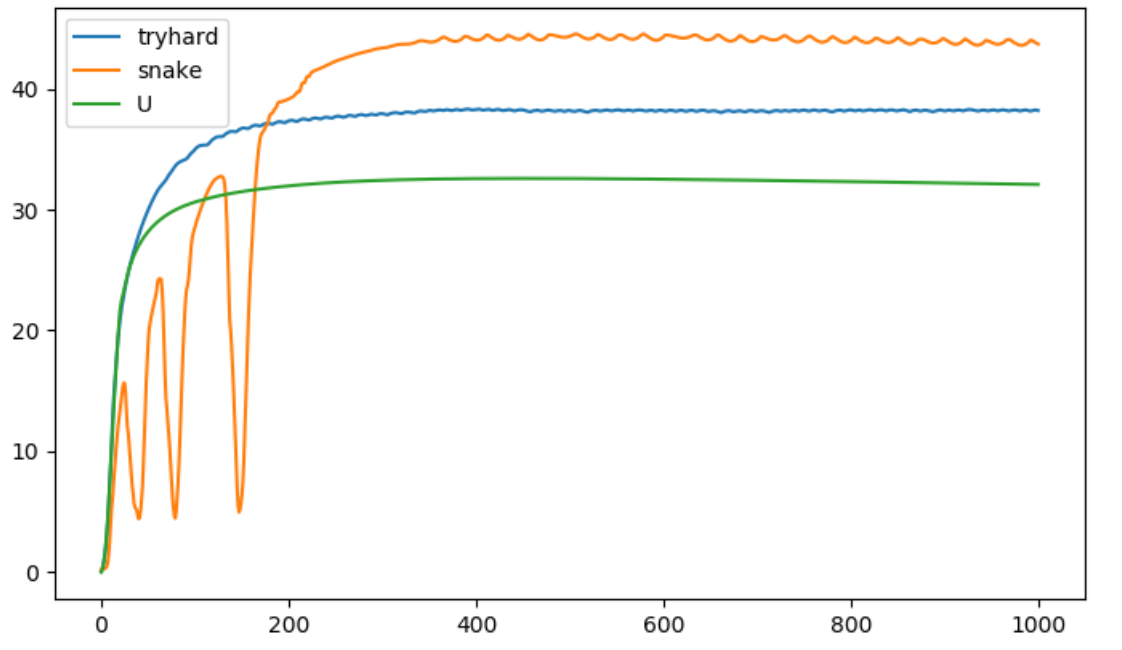}}
   \caption{Cumulative velocity weighted by $\gamma^t$ during an episode for different behaviors on \swimmer, with two different values of $\gamma$: (a) 0.99 and (b) 0.99999 \label{fig:swimmer_analysis}}
   \end{center}
\end{figure}

To better stress this point, in \figurename~\ref{fig:swimmer_analysis}, we depict the cumulative velocity weighted by $\gamma^t$ during an episode for different behaviors on \swimmer, with two different values of $\gamma$: 0.99 and 0.99999.

As the figure shows, with $\gamma = 0.99$ it is more optimal to throw the tail forward and then stay stuck than performing the snake-like behavior, as the greater initial velocity impulse matters more than the subsequent loss in velocity. By contrast, with $\gamma$ close enough to 1, the snake-like behavior provides a higher return.

Strikingly, we can say that, though the reported overall performance of \rl methods is suboptimal, this performance being the non-discounted sum of immediate rewards over 1000 time steps, in fact \rl methods do find the optimum behavior they are asked to optimize, that is the discounted performance.

By contrast with \rl methods, direct policy search methods do not consider immediate rewards separately, they just consider the global return as {\em fitness}, hence they do not use a discount factor $\gamma$. This is enough to explain that they ore easily converge to the snake-like behavior.

{\bf Thus, making \rl work on \swimmer is simple: one just as to set $\gamma=0.9999$ or $1$.}

\section{RL results with gamma = 0.9999}

\begin{figure}[!htbp]
  \begin{center}
{\includegraphics[width=0.99\linewidth]{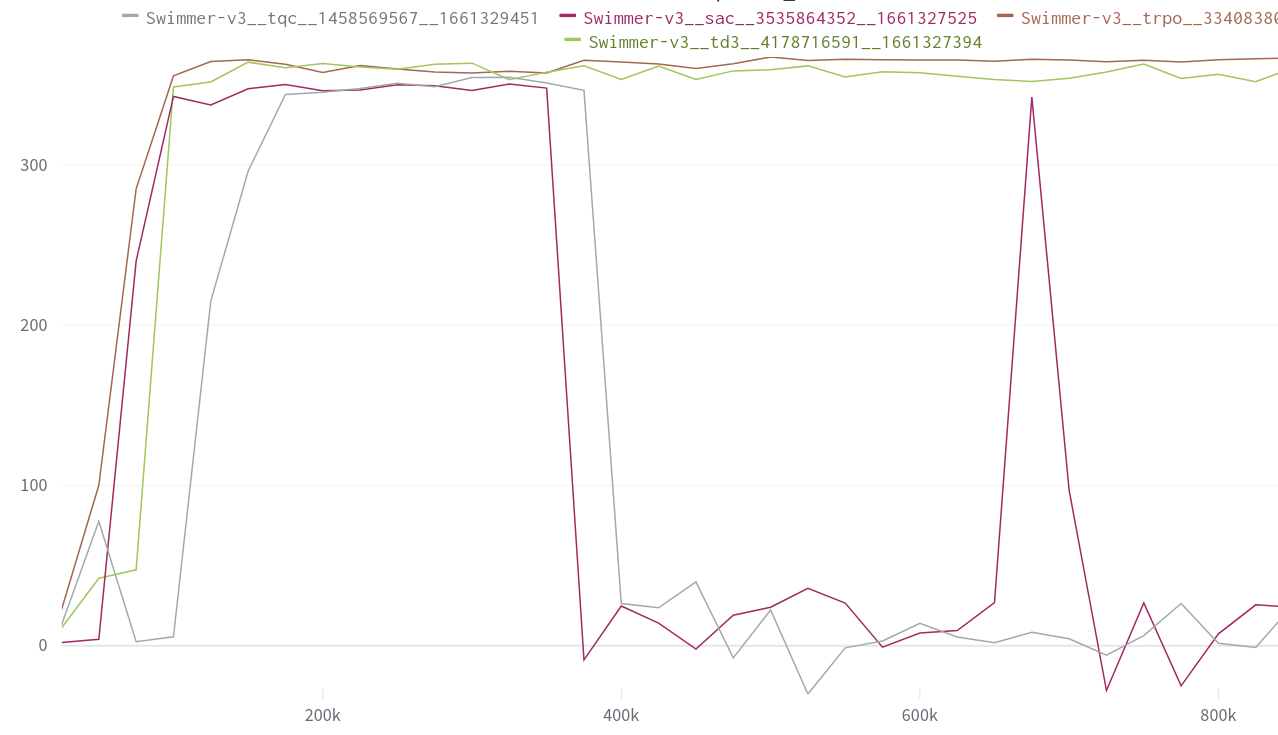}}
   \caption{
   Learning curves of \trpo, \tddd, \sac, and \tqc on the \swimmer environment.
   \label{fig:swimmer_rl}}
   \end{center}
\end{figure}

As \figurename~\ref{fig:swimmer_rl} shows, once properly tuned with $\gamma = 0.9999$, four standard \rl algorithms such as \trpo, \tddd \sac, and \tqc easily manage to reach a performance over 300, even if two of them show some instability. These results have been obtained with the Stable Baselines3 library \citep{raffin2021stable} and the RL Zoo \citep{raffin2020rlzoo}.

The hyper-parameters used to obtain the corresponding learning curves are available on the following pages: 
\begin{itemize}
\item
\href{https://huggingface.co/sb3/trpo-Swimmer-v3}{https://huggingface.co/sb3/trpo-Swimmer-v3}
\item
\href{https://huggingface.co/sb3/td3-Swimmer-v3}{https://huggingface.co/sb3/td3-Swimmer-v3}
\item
\href{https://huggingface.co/sb3/sac-Swimmer-v3}{https://huggingface.co/sb3/sac-Swimmer-v3}
\item
\href{https://huggingface.co/sb3/tqc-Swimmer-v3}{https://huggingface.co/sb3/tqc-Swimmer-v3}
\end{itemize}

Complete training log can be found in \href{https://wandb.ai/openrlbenchmark/sb3}{https://wandb.ai/openrlbenchmark/sb3}.

\section{Discussion}

A different fix to the above issue consists in moving the velocity sensor from the neck of the creature to its head. By doing this, the strategy consisting in throwing forward the tail does not impact much the velocity sensor, hence the local minimum issue that we have described in the previous section disappears. This other fix has been used for instance in \cite{wang2019exploring}, see Section A.2.1 of their paper.

\section{Conclusion}

In this paper, we have addressed a simple question: why is it that \rl methods do generally perform poorly on \swimmer, whereas direct policy search methods do not? Our analysis has resulted in a simple answer: the use of a discount factor in \rl methods results in inadequate evaluation of the generated behaviors, which is not the case in direct policy search methods. Setting the discount factor to 1 thus removes the issue. A consequence of our investigations is that, unfortunately, a lot of \rl results on \swimmer published in the last years must be considered with care or even reconsidered. Besides, we hope that the proposed analysis will prevent future authors from falling into the same trap.

\end{document}